\documentclass[runningheads]{llncs}

\usepackage{eccv}

\usepackage{eccvabbrv}

\usepackage{graphicx}
\usepackage{booktabs}
\usepackage{multirow}
\usepackage{seqsplit}
\usepackage{pifont}
\usepackage{colortbl}

\usepackage[accsupp]{axessibility}  % Improves PDF readability for those with disabilities.

\usepackage{array}    % 必须加载，用于定义M列格式
\usepackage{booktabs} % 专业表格线
\usepackage{makecell} % 用于\makecell命令
\usepackage{geometry} % 调整页面边距
\usepackage{float}
\usepackage[table]{xcolor} 
\usepackage[most]{tcolorbox}
\tcbuselibrary{breakable}
\usepackage{enumitem}
\usepackage{lmodern}
\usepackage[T1]{fontenc}
\usepackage{listings}
\usepackage{tabularx}

\newcolumntype{M}[1]{>{\centering\arraybackslash}m{#1}}

\usepackage{hyperref}

\usepackage{orcidlink}

\definecolor{LightBlue}{RGB}{235, 245, 252}    % 极浅蓝色（用于表头）
\definecolor{LightGray}{RGB}{250, 250, 250}    % 近乎白色的灰色（用于交替行）
\definecolor{DividerGray}{RGB}{240, 240, 240} % 稍微明显的灰色（用于分类栏）
\definecolor{OursHighlight}{RGB}{245, 240, 255} % 极浅绿色（用于突出 Ours）

\begin{document}

\title{Enhancing MLLM’s Spatial Understanding via Active 3D Scene Exploration for Multi-Perspective Reasoning} 


% \author{
% Jiahua Chen\inst{1}$^\ast$
% \and
% Qihong Tang\inst{2}$^\ast$
% \and
% Weinong Wang\inst{3}$^\dagger$
% \and
% Qi Fan\textsuperscript{\rm 2 \ding{41}}
% }
% \thanks{$^\ast$Equal contribution, $^\dagger$Project lead, \ding{41} Corresponding author}

% % TODO FINAL: Replace with an abbreviated list of authors.
% \authorrunning{J. Chen et al.}

% % TODO FINAL: Replace with your institution list.
% \institute{
% Tsinghua University \and Nanjing University \and Xi'an Jiaotong University\\
% \email{chenjiah25@mails.tsinghua.edu.cn}, \email{fanqi@nju.edu.cn}
% }
\author{Jiahua Chen\inst{1}\thanks{Equal contribution, $^\dagger$Project lead, \textsuperscript{\ding{41}}Corresponding author.}\and
Qihong Tang\inst{2}$^\star$\and
Weinong Wang\inst{3}$^\dagger$ \and
Qi Fan\inst{2}\textsuperscript{\ding{41}}}

\authorrunning{J. Chen et al.}
\titlerunning{Active 3D Exploration for MLLM Spatial Understanding}

\institute{$^1$ Tsinghua University, 
$^2$ Nanjing University,
$^3$ Tencent\\
\email{chenjiah25@mails.tsinghua.edu.cn, qhtang@smail.nju.edu.cn, weinong.wang@hotmail.com, fanqi@nju.edu.cn}}

\maketitle

\begin{abstract}
Although Multimodal Large Language Models have achieved remarkable progress, they still struggle with complex 3D spatial reasoning due to the reliance on 2D visual priors. 
Existing approaches typically mitigate this limitation either through computationally expensive post-training procedures on limited 3D datasets or through rigid tool-calling mechanisms that lack explicit geometric understanding and viewpoint flexibility. 
To address these challenges, we propose a \textit{training-free} framework that introduces a Visual Chain-of-Thought mechanism grounded in explicit 3D reconstruction. 
The proposed pipeline first reconstructs a high-fidelity 3D mesh from a single image using MLLM-guided keyword extraction and mask generation at multiple granularities. 
Subsequently, the framework leverages an external knowledge base to iteratively compute optimal camera extrinsic parameters and synthesize novel views, thereby emulating human perspective-taking. 
Extensive experiments demonstrate that the proposed approach significantly enhances spatial comprehension. 
Specifically, the framework outperforms specialized spatial models and general-purpose MLLMs, including \textit{GPT-5.2} and \textit{Gemini-2.5-Flash}, on major benchmarks such as 3DSRBench and Rel3D.
\end{abstract}

\section{Introduction}
\label{sec:intro}
Multimodal Large Language Models (MLLMs)~\cite{wang2025internvl3,bai2025qwen3,liu2024llava,li2024llava,bai2025qwen25vltechnicalreport,google2025gemini2.5,openai2026gpt52} have advanced rapidly, demonstrating impressive capabilities in visual and textual understanding. To further enable the deployment of MLLMs in downstream embodied applications in the real world~\cite{szot2025multimodal,mon2025embodied,qi2025vln,hong2021vln,lu2025vla,huangthinkact}—such as navigation~\cite{qi2025vln,hong2021vln} and robotic manipulation~\cite{lu2025vla,huangthinkact}—spatial understanding~\cite{wuspatial,ma2024spatialpin,yin2025spatial,chen2024spatialvlm,jia2026omnispatial,li2024topviewrs,guo2026thinking,helu2025scaling,ma2025spatialllm,liao2025improved} is a critical prerequisite. However, state-of-the-art commercial MLLMs still lag significantly behind human performance across nearly all existing spatial reasoning benchmarks~\cite{mirzaee2021spartqa,wu2025SpatialScore,goyal2020Rel3D,ma20253dsrbench,wang2025spatial457,lireframing}. This limitation stems primarily from the dominant pre-training paradigm. Most MLLMs are trained on aligned 2D image-text pairs, relying heavily on linguistic correlations and 2D visual priors while lacking a genuine understanding of 3D structures and spatial relations~\cite{hu2025g2vlmgeometrygroundedvision}. Current models often fail to achieve precise object-level perception in 3D space, as well as to infer the spatial configuration and the underlying causal and physical structure of real-world environments~\cite{wang2025n3d}. Therefore, advancing toward genuine multimodal intelligence requires a transition from 2D image-text-centric perception toward comprehensive spatial capabilities that facilitate perception, construction, and reasoning within the 3D world.

\begin{figure}[!t]
\centering
\includegraphics[width=1\textwidth]{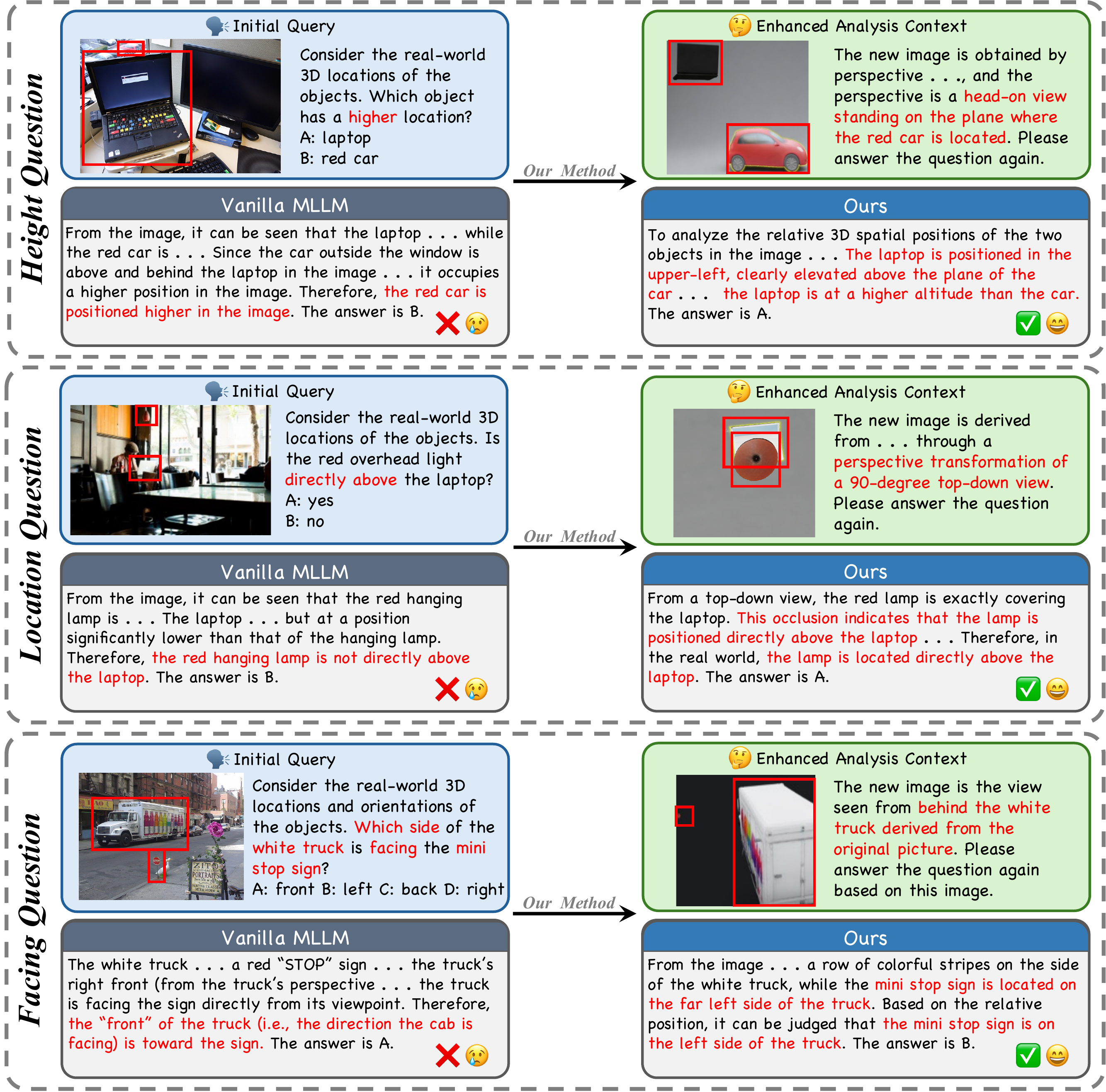}
\caption{Effectiveness of the proposed Visual CoT method driven by 3D reconstruction. Vanilla MLLMs (left) often fail to answer complex spatial questions correctly, including height comparisons, location assessments, and facing orientations, due to perspective ambiguities and reliance on 2D visual priors derived from a single static image. In contrast, the proposed approach (right) reconstructs the underlying 3D scene and applies perspective transformations to synthesize novel and geometrically unambiguous viewpoints (\eg, head-on, top-down, or specific relative views). By dynamically transitioning to these informative perspectives, the proposed method resolves spatial ambiguities and enables accurate 3D spatial reasoning.}
\label{fig:teaser}
\end{figure}

Mainstream approaches designed to enhance 3D spatial understanding in MLLMs inject geometric and depth features through explicit 3D integration frameworks~\cite{hu2025g2vlmgeometrygroundedvision, chen2025geometrically, chen2025think}, which partially mitigate the dimensional constraints. However, such paradigms still suffer from limitations. Several methods represent 3D scenes using abstract numerical values or shapes~\cite{ma2024spatialpin, chen2026cvp}. This representation not only discards geometric details but also deviates from the natural process of human spatial reasoning through vision. Other methods~\cite{lee2025perspective, cao2025spatialdreamer} attempt to reason from an egocentric view for the assessment of spatial relations. Yet, the limited first-person perspective often leads to incomplete observations and erroneous answers when compared with third-person views, thereby restricting the range of applicable tasks. Furthermore, despite evidence from autonomous driving that multi-view perception improves scene understanding based on MLLMs~\cite{chen2019point, yu2020fast, yang2022mvs2d}, few methods leverage multi-view inputs to enhance the comprehension of spatial relationships.

To address these issues, we draw inspiration from human spatial cognition. When assessing spatial relationships, humans typically resolve ambiguities by mentally reconstructing scenes and imagining the environment from different viewpoints. We find that these viewpoints also assist MLLMs in understanding spatial relations, as illustrated in~\cref{fig:teaser}. We formalize this strategy as a Visual Chain-of-Thought (Visual CoT) mechanism driven by 3D reconstruction. Analogous to textual Chain-of-Thought (CoT), which decomposes logical reasoning into intermediate linguistic steps, the proposed visual mechanism explicitly models the process of spatial reasoning. By actively synthesizing novel viewpoints, the model explores the space of camera poses until it identifies geometrically unambiguous perspectives. During this visual exploration, the model continuously captures multiple views of the scene, thereby enriching scene comprehension and improving the reasoning of spatial relations. Specifically, guided by the MLLM, the system first extracts keywords of multiple granularities from the input query. These keywords are subsequently used to generate masks, ensuring complete scene modeling for the subsequent reconstruction of the 3D mesh. In the second stage, the framework performs dynamic view transitions. The MLLM infers the optimal observation perspective for the given task by querying a spatial knowledge base. The framework computes the extrinsic parameters of the camera and transforms spatial coordinates into the world coordinate system to synthesize novel views. Additionally, we introduce an iterative selection mechanism to ensure high robustness. This closed-loop verification process continuously optimizes the observation space.

We evaluate the proposed framework across spatial benchmarks, including 3DSRBench~\cite{ma20253dsrbench}, SpatialScore~\cite{wu2025SpatialScore}, and Rel3D~\cite{goyal2020Rel3D}. The proposed approach outperforms both specialized spatial models and general-purpose models. On 3DSRBench and SpatialScore, the proposed method achieves average accuracy improvements of $5.6\%$ and $4.1\%$, respectively, compared with the state-of-the-art model \textit{Gemini-2.5-Flash}~\cite{google2025gemini2.5}. This advantage is particularly notable in the zero-shot setting, where the method achieves an average accuracy improvement of $1.0\%$ on Rel3D compared with \textit{GPT-5.2}~\cite{openai2026gpt52}, which is one of the most capable closed-source models.

In summary, this work makes three main contributions:
\begin{itemize}
\item We propose a \textit{training-free} framework that fuses 3D scene reconstruction and the Visual CoT mechanism, thereby achieving high robustness.
\item We enable MLLMs to freely explore 3D space and actively perform iterative view transitions to obtain informative viewpoints and enhance scene cognition, which strengthens the spatial reasoning capabilities of the models.
\item We achieve \textit{state-of-the-art} results on multiple benchmarks for spatial understanding, demonstrating measurable improvements over existing architectures.
\end{itemize}

\section{Related Works}

\noindent\textbf{Spatial reasoning. } Spatial reasoning is the ability to perceive and infer geometric and logical relationships among objects in three-dimensional (3D) space. 

Efforts improving MLLM spatial reasoning~\cite{wuspatial,ma2024spatialpin,yin2025spatial,chen2024spatialvlm,jia2026omnispatial,li2024topviewrs,guo2026thinking,helu2025scaling,ma2025spatialllm,liao2025improved} generally follow two paradigms. 
Test-time scaling uses methods such as Retrieval-Augmented Generation (RAG)~\cite{lewis2020retrieval,jiang2023active}, prompt engineering~\cite{wei2022chain}, and tool calling~\cite{yao2022react}. 
Prompt engineering elicits spatial reasoning via explicit instructions. 
While approaches like TopViewRS~\cite{li2024topviewrs} use simple templates, the resulting performance gains are often limited. 
OmniSpatial~\cite{jia2026omnispatial} points out the constraints of text-based CoT~\cite{wei2022chain} reasoning, suggesting that hybrid-modal prompt frameworks offer a better direction. 
Alternatively, tool calling lets models act as agents accessing external models, code, or tools, improving spatial reasoning without architectural changes. 
For example, CodeVision~\cite{guo2026thinking} helps MLLMs process images and videos by calling external tools.
However, these techniques rely heavily on comprehensive tool libraries, and building flexible, scalable toolkits remains a challenge.
Post-training approaches enhance MLLMs via SFT and RL~\cite{schulman2017proximal,shao2024deepseekmath}. 
SPRITE~\cite{helu2025scaling} trains MLLMs on programmatically synthesized data to improve performance on spatial benchmarks. 
However, reliance on high-quality spatial datasets bottlenecks SFT methods due to data scarcity. 
vsGRPO~\cite{liao2025improved} uses a self-play paradigm to reduce this dependency by automatically generating and solving spatial problems, using training strategies similar to those in R1-Zero~\cite{guo2025deepseek,huopen,zhou2025r1}. 
While these RL methods complement SFT, they require careful reward design and introduce high computational overhead. 
Unlike these paradigms, our method \textit{requires no training}. 
It improves spatial understanding by directly prompting 3D reconstruction models, offering a simple and efficient alternative to computationally expensive post-training pipelines.

\noindent\textbf{Visual Chain-of-Thought. }
Chain-of-Thought~\cite{wei2022chain} reasoning has proven highly effective in improving the problem-solving capabilities of large language models (LLMs) by prompting them to generate intermediate steps. 
This strategy naturally extends to MLLMs~\cite{zheng2023ddcot,hu2024visual,yang2023set,zhou2024image,shao2024visual,dong2025seeing,zhao2025cot,lin2025showui}. 
Rather than relying strictly on textual intermediate steps, recent methods incorporate visual features directly into the prompting process to improve image reasoning. 
ShowUI~\cite{lin2025showui} demonstrates this by training an MLLM to process images augmented with bounding boxes. 
Other approaches provide models with explicit visual annotations. 
Visual Sketchpad~\cite{hu2024visual} allows vision-language models to annotate inputs using drawing tools for tasks such as mathematical problem-solving, while Set-of-Mark~\cite{yang2023set} overlays segmentation masks on objects to provide fine-grained visual context. 
To better analyze MLLM cognitive processes, IoT~\cite{zhou2024image} automatically formulates operations extracting key visual information based on the specific input image and question. 
Motivated by these approaches, we introduce a Visual CoT mechanism into the reasoning process specifically to improve the spatial reasoning capabilities of MLLMs.

\noindent\textbf{MLLMs with 3D Reconstruction. }
Integrating 3D models to supply geometric features helps learn visual geometry~\cite{chen2024spatialvlm,fan2025vlm3rvisionlanguagemodelsaugmented,hu2025g2vlmgeometrygroundedvision,guo2025beyond,lee2025perspective,ma2024spatialpin}. For instance, G$^2$VLM~\cite{hu2025g2vlmgeometrygroundedvision} uses dedicated geometry- and semantic-aware experts to learn 3D geometry from 2D data, improving spatial reasoning through shared self-attention to achieve state-of-the-art results on several benchmarks. 
GEODE~\cite{guo2025beyond} aligns explicit 3D data with 2D visual features via cross-attention to maintain strong performance at a smaller model scale. 
While effective, incorporating this heterogeneous information usually requires extensive training data and high computational overhead.
Other approaches avoid heavy training pipelines. 
SpatialPin~\cite{ma2024spatialpin} offers a modular, training-free framework that uses progressive prompting to decompose and reconstruct explicit 3D representations. 
APC~\cite{lee2025perspective} models spatial understanding through a 3D scene abstraction, but it causes significant information loss and restricts the method to specific tasks.

To resolve these issues, our proposed method introduces a 3D reconstruction model that enables MLLMs to observe constructed scenes from arbitrary perspectives. 
By alternating between linguistic and visual reasoning to approximate human cognition, our method improves performance on a broader range of spatial understanding tasks.

\begin{figure}[!t] %H为当前位置，!htb为忽略美学标准，htbp为浮动图形
\centering %图片居中
\includegraphics[width=\textwidth]{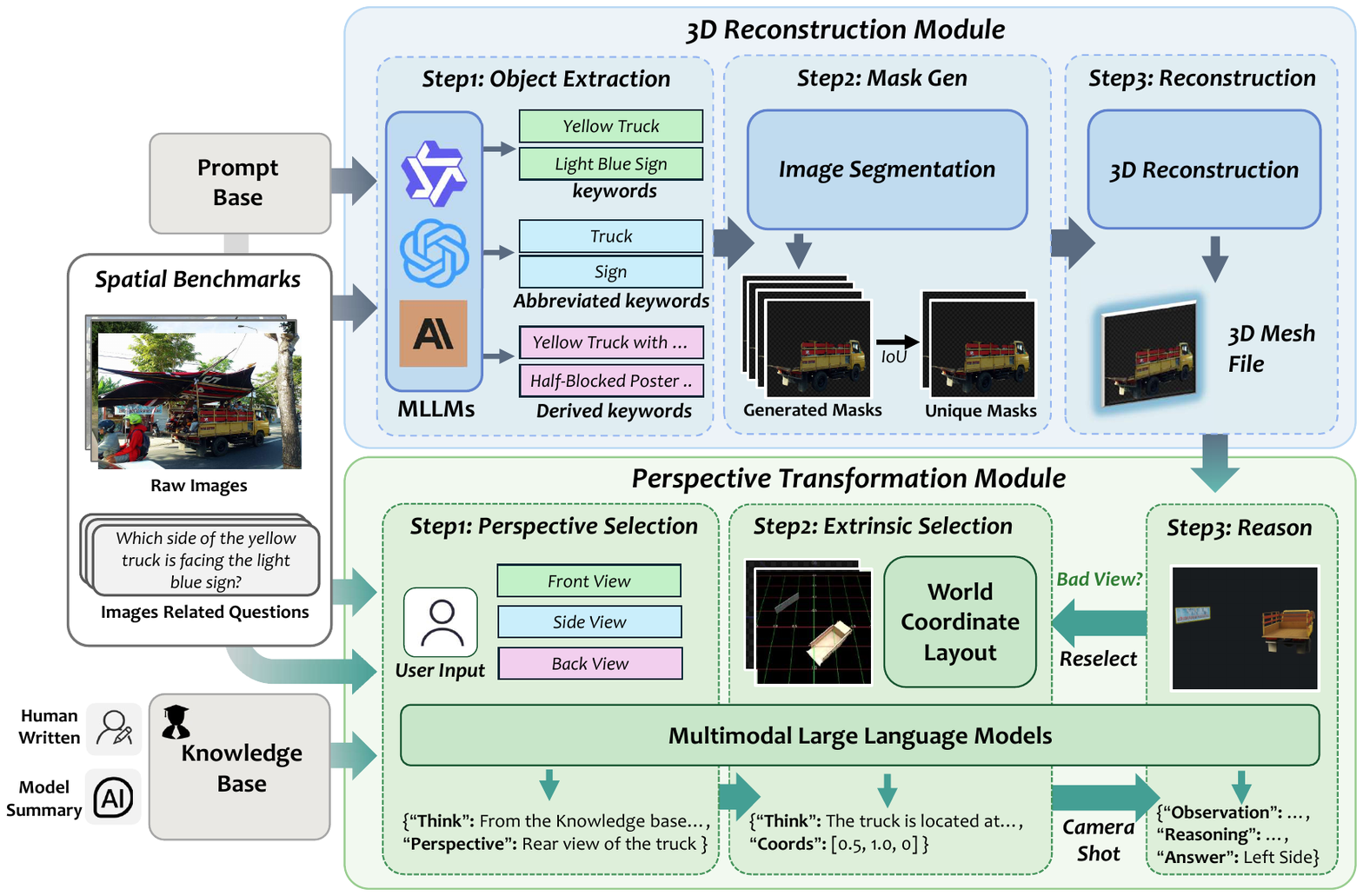}
\caption{Architecture of the proposed end-to-end framework for spatial comprehension, comprising the \textit{3D reconstruction} and \textit{perspective transformation} modules. The reconstruction phase employs MLLMs to extract multi-granularity keywords, guiding mask generation and IoU deduplication to construct a three-dimensional mesh. The perspective transformation phase integrates an external knowledge base for optimal viewpoint selection, computes the extrinsic parameters of the camera within a unified world coordinate system, and executes an iterative view synthesis loop with a re-selection mechanism to generate the final reasoning response.} %最终文档中希望显示的图片标题
\label{fig:pipeline} %用于文内引用的标签
\end{figure}%结束环境

\section{Methodology}

To address the limited spatial perception of 2D images and the performance degradation of MLLMs caused by the restricted field of view in single images, we propose a pipeline that enables MLLMs to accurately answer spatial comprehension questions about visual content. 
This section provides an overview of the end-to-end workflow, from data processing to model reasoning, followed by implementation details for the data processing and perspective transformation stages.

\noindent\textbf{Overview. } 
As shown in \cref{fig:pipeline}, the proposed framework processes the raw images and the corresponding spatial comprehension questions through two main components. 
In the \textit{3D reconstruction module}, object-centric prompts from a predefined library guide an MLLM (specifically, Qwen3-VL-235B-A22B~\cite{bai2025qwen3}) to describe the input image and the associated question. 
These descriptions capture the object attributes (\eg, appearance) and the spatial relationships. 
The MLLM subsequently extracts \textit{original, abbreviated, and derived keywords} based on the generated descriptions. 
This extraction at multiple granularities assists the subsequent segmentation model in locating the relevant target objects.
Following the extraction of the keywords, the segmentation model generates the initial object masks. 
These masks are deduplicated using the Intersection-over-Union (IoU) metric to remove redundancy, yielding a unique and high-fidelity set of masks. 
The raw image and the refined masks are then provided to a 3D reconstruction model to generate a mesh representation of the key objects. 
This mesh representation enables the arbitrary manipulation of the camera pose within the 3D space, which is essential for the synthesis of images from novel viewpoints.

The \textit{perspective transformation module} relies on a knowledge base compiled from human annotations and learned perspective rules, integrated directly into the MLLM (specifically, Qwen3-VL-8B-Instruct~\cite{bai2025qwen3}). 
By utilizing this knowledge base alongside the original image and the question from the dataset, the MLLM determines an optimal perspective description to formulate the answer. 
A unified world coordinate system is then established for all mesh objects by translating the origin to the geometric center of the objects. 
A two-dimensional Cartesian system defines the horizontal plane, while a spherical coordinate system models the pitch angles of the camera on the vertical plane. 
Based on this coordinate framework, the MLLM predicts the ideal position of the camera. 
The extrinsic parameters of the camera are computed from this position, whereas the intrinsic parameters are defined a priori. 
The 3D mesh is subsequently rendered to synthesize the novel view, which is returned to the MLLM as a visual description. 
The model compares this synthesized view against the initial perspective description. 
If the synthesized view is inadequate (\eg, due to occlusion, incompleteness, or misalignment), a re-selection mechanism prompts the MLLM to estimate a new pose for the camera. 
This iterative refinement continues until the view is validated, thereby allowing the MLLM to generate the final answer to the original question.

\subsection{3D Reconstruction Module}
\label{sec:3d_recon}

This module extracts target objects from the input image and the associated spatial comprehension question to reconstruct their corresponding 3D meshes. 
The process consists of three steps: \textit{object extraction}, \textit{mask generation}, and \textit{3D reconstruction}.

\noindent\textbf{Step 1: Object Extraction.}
The image and the question are processed with specific prompts to extract key object terms. 
We use Qwen3-VL-235B-A22B~\cite{bai2025qwen3} for keyword extraction due to its strong visual reasoning and text expansion capabilities. 
Empirically, object terms derived directly from the question are often insufficient for the segmentation model to produce accurate masks.
To address this, we instruct the model to generate keywords at three levels of granularity:

\begin{itemize}
    \item \textbf{Original keywords}: Object terms directly extracted from the question.
    \item \textbf{Expanded keywords}: Keywords indicating location (\eg, relative position to the image or other objects), appearance adjectives, and synonyms.
    \item \textbf{Abbreviated keywords}: The original keywords stripped of modifiers, retaining only the core noun.
\end{itemize}

Inputting these three keyword variants into the segmentation model significantly increases the number of valid masks. 
% In rare cases where the MLLM fails to identify key objects or generates noisy outputs, we resolve the exceptions through manual annotation and filtering.

\noindent\textbf{Step 2: Mask Generation.}
The extracted keywords are fed into an image segmentation model to generate masks, which are then deduplicated to retain one mask per object. 
We employ SAM3~\cite{carion2025sam3} for this task. 
As a Promptable Concept Segmentation (PCS) model, it takes the raw image and text prompts to generate masks for specific concepts, aligning well with the MLLM-generated keywords. 
Since redundant keywords often yield overlapping masks, we apply an IoU deduplication process. 
The pairwise IoU score for all generated masks is computed as follows:
\begin{equation}
    \text{OverlapScore}_{ij} = \text{IoU}(m_i, m_j) = \frac{\text{Cnt}(m_i \cap m_j)}{\text{Cnt}(m_i \cup m_j)}
    \label{eq:overlap_score}
\end{equation}

where $\text{Cnt}$ denotes the number of true elements in the matrix. 
If the IoU score exceeds a predefined threshold of 0.7, we retain only the mask with the smaller area.

\noindent\textbf{Step 3: 3D Reconstruction.}
Finally, the generated masks and the raw image are processed by the SAM3D-Object~\cite{chen2025sam3d} model to produce the mesh file. 
This model supports the 3D completion of occluded objects, enabling complete modeling of the scene. 
Through parameter tuning, we ensure the layout of the reconstructed objects aligns well with their original arrangement in the input image, minimizing error propagation to subsequent pipeline stages.

\subsection{Perspective Transformation Module}
\label{sec:persp_trans}
This module performs viewpoint transformation to assist in answering complex spatial reasoning questions using an external \textit{knowledge base}. 
The process consists of three primary steps: \textit{perspective selection}, \textit{camera extrinsic prediction}, and \textit{view synthesis}.

\noindent\textbf{Knowledge Base.}
Vanilla MLLMs often struggle to select optimal viewpoints for specific spatial comprehension tasks. 
Inspired by RAG, we introduce an external knowledge base to improve the model's reasoning capabilities. 
When encountering similar problems, the model can refer to viewpoints that successfully solved prior questions or extrapolate effective viewing strategies from existing reasoning patterns. 
To construct this knowledge base, we select 6 out of 12 sub-tasks from the 3DSR~\cite{ma20253dsrbench} benchmark and sample a subset, yielding a reference set accounting for only 1\% of the full dataset.
We render comprehensive multi-view images for each sample and query the MLLM to answer the corresponding spatial questions. 
We then filter the correctly answered dialogues, summarize them using the MLLM, and refine the summaries through human curation to obtain task-related knowledge items. 
When addressing novel questions, the MLLM can infer useful viewpoints by generalizing from the logic encoded within the knowledge base. 
We validate this generalization ability in \cref{sec:case_study}.

\noindent\textbf{Step 1: Perspective Selection.}
We provide the MLLM with the constructed knowledge base, the input image, and the dataset question. 
The MLLM identifies the task category, selects an optimal viewpoint, and generates a linguistic description of that viewpoint.

\noindent\textbf{Step 2: Camera Extrinsic Prediction.}
We compute the camera extrinsic parameters that align with the textual description of the target viewpoint. 
This computation relies on the world coordinate system of the 3D mesh file to facilitate the rendering of the required viewpoint images.

We first calculate the geometric center $\boldsymbol{C}$ of the 3D bounding box containing all objects. Assuming $N$ objects are extracted from the scene, the vertex coordinates for the $i$-th object are denoted as $\mathcal{V}_i = \{v_{ij},\ j=1,2,\dots\}$, where $i \in \{1,2,\dots,N\}$. For each object, we compute the 3D bounding box from $\mathcal{V}_i$ by finding the minimum and maximum vertex values along the $x$, $y$, and $z$ axes. This yields the diagonal points of the bounding box, $\boldsymbol{b_{\min}}$ and $\boldsymbol{b_{\max}}$:
\begin{equation}
\begin{aligned}
\boldsymbol{b_{\min}} &= \left[ \min_x v_{ij}(x),\ \min_y v_{ij}(y),\ \min_z v_{ij}(z) \right], \\
\boldsymbol{b_{\max}} &= \left[ \max_x v_{ij}(x),\ \max_y v_{ij}(y),\ \max_z v_{ij}(z) \right].
\end{aligned}
\end{equation}

The geometric center $\boldsymbol{C}$ of the entire scene is then computed as the midpoint of these diagonal points:
\begin{equation}
    \boldsymbol{C} = \frac{\boldsymbol{b_{\min}} + \boldsymbol{b_{\max}}}{2}
\end{equation}

We use $\boldsymbol{C}$ as the origin of the object coordinate system. To ensure all target objects remain fully observable, the camera consistently faces this center. 
The $x$ and $y$ axes of this object coordinate system align with the $x$ and $y$ directions of the world coordinate system. 
The $z$-axis remains aligned with the world $z$-axis ($\boldsymbol{n} = [0, 0, 1]$). 
This transformation converts the vertex coordinates of each object into the new object coordinate system.

Because 2D images primarily convey information on a 2D plane, we initially prompt the MLLM to predict the camera position $(\hat{x}', \hat{y}')$ on the $xy$ plane, starting from a top-down perspective. 
Based on this planar position, we calculate the center distance $\hat{r}$ and the yaw angle $\widehat{yaw}$:
\begin{equation}
\begin{cases}
\hat{r} = \sqrt{\hat{x}'^2 + \hat{y}'^2} \\
\widehat{yaw} = \arctan2(\hat{y}', \hat{x}')
\end{cases}
\end{equation}
where $\hat{r}$ denotes the distance from the camera to the object center $\boldsymbol{C}$, and $\widehat{yaw}$ represents the horizontal rotation angle of the camera within the $xy$ plane. 
We use the $\arctan2(y,x)$ function rather than $\arctan(\frac{y}{x})$ to correctly distinguish the angle's quadrant, yielding a continuous value in the range $(-\pi, \pi]$.

We then shift to a side view and construct a spherical coordinate system that enables $\pm 90^\circ$ vertical viewing range, and encompasses only the pitch angle dimension, allowing the MLLM to predict the pitch angle $\widehat{pitch}$ based on task requirements. 
After obtaining $\widehat{yaw}$, $\widehat{pitch}$, and $\hat{r}$, we calculate the camera position $\boldsymbol{P'}$ in the object coordinate system, which is then converted to the absolute position $\boldsymbol{P}$ in the world coordinate system:
\begin{equation}
\boldsymbol{P'} = \left[ \cos(\widehat{yaw}) \cdot \cos(\widehat{pitch}),\ \sin(\widehat{yaw}) \cdot \cos(\widehat{pitch}),\ \sin(\widehat{pitch}) \right] \cdot \hat{r}
\end{equation}

Here, $\boldsymbol{P'}$ represents the camera position relative to the origin $\boldsymbol{C}$. Adding $\boldsymbol{C}$ translates $\boldsymbol{P'}$ back to the absolute position $\boldsymbol{P}$ in the world coordinate system.
\begin{equation}
\boldsymbol{P} = \boldsymbol{P'} + \boldsymbol{C} 
\end{equation}

Finally, we compute the camera extrinsic matrix $\boldsymbol{E}$ based on position $\boldsymbol{P}$ and the constraint that the camera faces the center $\boldsymbol{C}$. 
The extrinsic matrix consists of a rotation matrix $\boldsymbol{R_{3\times3}}$ and a translation vector $\boldsymbol{t_{3\times1}}$:
\begin{equation}
\boldsymbol{E} = \begin{bmatrix}
\boldsymbol{R}_{3\times3} & \boldsymbol{t}_{3\times1} \\
\boldsymbol{0}_{1\times3} & 1
\end{bmatrix},
\end{equation}

The three columns $\boldsymbol{Rx}$, $\boldsymbol{Ry}$, and $\boldsymbol{Rz}$ of the rotation matrix $\boldsymbol{R_{3\times3}}$ are obtained through vector cross products and normalization to guarantee orthogonality:
\begin{equation}
\begin{aligned}
     \boldsymbol{R}_{3\times3} =& \left[ \boldsymbol{Rx}_{3\times1},\ \boldsymbol{Ry}_{3\times1},\ \boldsymbol{Rz}_{3\times1} \right] \\
     =& \left[ \text{Norm}(-\boldsymbol{n} \times \boldsymbol{Rz}),\ \text{Norm}(\boldsymbol{Rz} \times \boldsymbol{Rx}),\ \text{Norm}(-\boldsymbol{P'}) \right]   
\end{aligned}
\end{equation}

The translation vector $\boldsymbol{t_{3\times1}}$ is calculated from the rotation matrix and the camera position $\boldsymbol{P}$:
\begin{equation}
    \boldsymbol{t}_{3\times1} = -\boldsymbol{R P}
\end{equation}

This process yields the camera extrinsic matrix $E$ aligned with the target viewpoint, which is subsequently used for novel view rendering.

\noindent\textbf{Step 3: View Synthesis and Reasoning.}
We synthesize the novel view using the predicted extrinsic parameters and the predefined camera intrinsics. 
The MLLM is prompted to describe the rendered view and to verify the consistency of the view with the viewpoint description generated in the first stage. 
If the synthesized view exhibits defects, such as incomplete objects, improper scaling, severe occlusion, or inconsistencies with the target description, the pipeline reverts to the previous step to regenerate the camera parameters. 
This iterative refinement process continues until the MLLM validates the rendered view or a predefined retry limit is reached. Before  final question answering, we prompt the model to describe each synthesized novel-view image, allowing the MLLM to incorporate and reason from multi-view visual cues for better scene understanding.

\section{Experiments}
\label{sec:exp}

\subsection{Experimental Settings}
\label{sec:exp_setting}

\begin{figure}[!t]
    \centering
    \includegraphics[width=0.7\columnwidth]{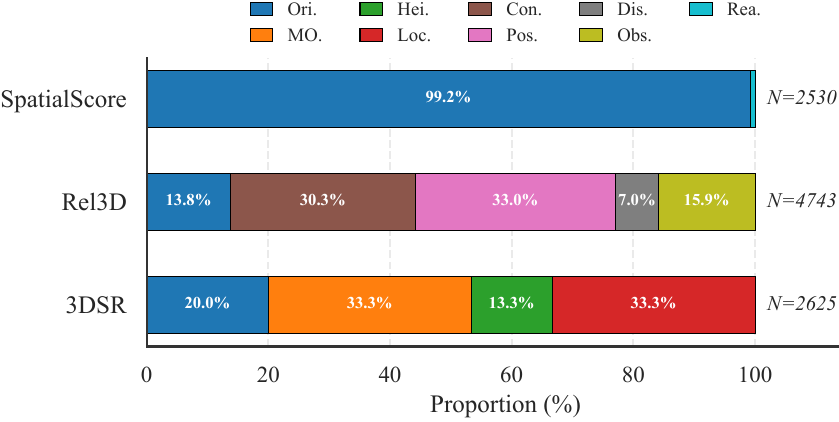}
    \caption{Distribution of sub-task types across different benchmarks (3DSR, Rel3D, and SpatialScore). The horizontal axis indicates the relative proportion of each task category, whereas the absolute number of samples ($N$) for each benchmark is annotated on the right. A significant bias towards the Orientation task is observable in the SpatialScore dataset. The abbreviations for the sub-tasks denote the following: Hei. (Height), Loc. (Location), Ori. (Orientation), MO. (Multi-Object), Rea. (Reach), Con. (Containment), Pos. (Position), Dis. (Distance), and Obs. (Observer).}
    \label{fig:benchmark_distribution}
\end{figure}

\noindent\textbf{Benchmarks. }
To evaluate the spatial reasoning capabilities of the proposed method, we utilize three benchmarks: 3DSRBench~\cite{ma20253dsrbench}, Rel3D~\cite{goyal2020Rel3D}, and SpatialScore~\cite{wu2025SpatialScore}. The data volume of each benchmark, along with the subtasks constituting the respective task types and the corresponding proportions, is illustrated in \cref{fig:benchmark_distribution}. All subtasks within these benchmarks consist of diverse question types; the specific question types corresponding to these subtasks are detailed in the supplementary material.

\begin{itemize}
    \item \textbf{3DSRBench} assesses 3D spatial perception within natural images. It contains 2,772 human-annotated visual question answering pairs covering 12 question types (\eg, height, orientation, and multi-object reasoning). Following the original publication, we categorize the subtasks into four primary tasks and remove samples for duplication and knowledge base generation.
    \item \textbf{Rel3D} evaluates the localization of spatial relationships within 3D environments through minimal contrast pairs, encompassing up to 30 distinct spatial relations. For each scene, a specific spatial relationship holds in one scenario but does not hold in the corresponding counterpart. These 30 relations were processed using a large language model to generate annotations, yielding five broad categories. To increase dataset complexity, we modified the options for samples lacking the spatial relationship into a binary yes/no format. For samples where the spatial relationship is present, we constructed two or four multiple-choice options. One option serves as the ground truth, while the remaining options represent alternative spatial relations within the same category.
    \item \textbf{SpatialScore} provides a comprehensive evaluation of multimodal spatial intelligence. This benchmark comprises approximately 5,000 human-validated samples across 30 tasks, assessing the performance of the model across various visual data types, input modalities, and QA formats. To evaluate the proposed method on this dataset, we filtered the samples to isolate the image modality and exclusively selected tasks related to the assessment of 3D spatial relationships between objects.
\end{itemize}

% We evaluate the spatial reasoning capabilities of our method using three benchmarks: 3DSRBench~\cite{ma20253dsrbench}, Rel3D~\cite{goyal2020Rel3D}, and SpatialScore~\cite{wu2025SpatialScore}. 
% 3DSRBench evaluates 3D spatial awareness in natural images and contains 2,772 manually annotated visual question answering (VQA) pairs spanning 12 question types (\textit{e.g.}, height, orientation, and multi-object reasoning). 
% % We use its paired image subsets from both common and uncommon 6D viewpoints, combined with the FlipEval strategy, to measure model robustness against camera viewpoint variations. 
% Rel3D evaluates the grounding of spatial relations in 3D environments, which provides minimally contrastive pairs, consisting of nearly identical 3D scenes where a spatial relation holds in one but fails in the other. 
% This structure helps mitigate dataset bias and provides a precise diagnostic evaluation of relation detection. 
% SpatialScore offers a broader evaluation of multimodal spatial intelligence. 
% With approximately 5,000 manually verified samples across 30 tasks, it evaluates the model on various visual data types, input modalities, and QA formats, providing a comparison with current state-of-the-art MLLMs.

\noindent\textbf{Baselines. }
We evaluate the proposed model against recent state-of-the-art MLLMs, which are categorized into \textit{general-purpose models} and \textit{spatially specialized models}. 
The general-purpose baselines exhibit standard visual-language understanding capabilities across diverse architectures and parameter scales. 
The evaluated \textit{closed-source} models include GPT-5.2~\cite{openai2026gpt52} and Gemini-2.5-Flash~\cite{google2025gemini2.5}. 
For \textit{open-source} comparisons, we evaluate Qwen3-VL-8B-Instruct~\cite{bai2025qwen3}, InternVL3.5-8B~\cite{wang2025internvl3}, LLaVA-OneVision-1.5-8B-Instruct~\cite{an2025llava}, LLaVA-v1.6-Mistral-7B~\cite{liu2024llava}, and Cambrian-1-8B~\cite{tong2024cambrian}.
To assess spatial and 3D structural reasoning, we compare the proposed method against specialized models. 
SpaceMantis-8B~\cite{remyx2025spacemantis} and SpaceQwen2.5-VL-3B~\cite{remyxai2025spaceqwen25vl} enhance spatial relationship reasoning in visual question answering via data synthesis and model ensembling. 
Spatial-MLLM-7B~\cite{wuspatial} utilizes a dual-encoder architecture to extract structural priors from visual geometry foundation models, facilitating 2D-to-3D reasoning. 
VLM3R-7B~\cite{fan2025vlm3rvisionlanguagemodelsaugmented} incorporates a geometry encoder and instruction tuning to process spatial-temporal information derived from monocular videos. 
Furthermore, we evaluate SpatialLLM-8B~\cite{ma2025spatialllm}, which is fine-tuned on 3D-informed conversations to process location and orientation relationships.

\noindent\textbf{Setup. } 
We use Qwen3-VL-8B-Instruct~\cite{bai2025qwen3} as the base model. 
All parameters for text generation are set to default. 
During inference, the model outputs responses in \texttt{JSON} format to ensure reliable information extraction for subsequent steps. 

\begin{table}[t]
\centering
\caption{Performance comparison of various models on the 3DSR, SpatialScore, and Rel3D benchmarks, categorized by model type. The symbol $\ddagger$ denotes models that exclusively process video inputs; during evaluation, static images are formatted as single-frame videos for these baselines. $\star$ denotes a state-of-the-art model with $\sim$100B active parameters (total estimated larger), trained on $\ge$30T tokens with an estimated $10^{25}$–$10^{26}$ FLOPs. Abbreviations follow the definitions provided in \cref{fig:benchmark_distribution}. The best and second-best results for each metric are indicated in \textbf{bold} and \underline{underlined} text, respectively. All results denote accuracy (\%).}
\label{tab:main}
\resizebox{\linewidth}{!}{
\begin{tabular}{l|ccccc|ccc|cccccc}
\toprule
\rowcolor{LightBlue}
 & \multicolumn{5}{c}{\textbf{3DSR}} & \multicolumn{3}{c}{\textbf{SpatialScore}} & \multicolumn{6}{c}{\textbf{Rel3D}} \\
% \cmidrule(lr){2-6} \cmidrule(lr){7-9} \cmidrule(lr){10-15}
\rowcolor{LightBlue}
\multirow{-2}{*}{\textbf{Method}} & Hei. & Loc. & Ori. & MO. & \textbf{Avg.} & Ori. & Rea. & \textbf{Avg.} & Con. & Pos. & Dis. & Obs. & Ori. & \textbf{Avg.} \\
\midrule
\rowcolor{DividerGray}
\multicolumn{15}{l}{\textit{Open-source Models}} \\
LLaVa-1.6-Mistral-7B~\cite{liu2024llava} & 51.1 & 57.5 & 40.2 & 46.3 & 49.5 & 60.7 & \underline{68.4} & 60.8 & 57.4 & 41.3 & \underline{57.8} & 34.5 & 34.3 & 45.3 \\
Cambrian-1-8B~\cite{tong2024cambrian} & 54.9 & 64.2 & 43.4 & 52.3 & 54.9 & 62.6 & 57.9 & 62.6 & 53.7 & 34.6 & 41.9 & 34.8 & 35.2 & 41.0 \\
LLaVA-OneVision-1.5-8B~\cite{an2025llava} & 52.9 & 60.5 & 43.2 & 47.4 & 51.7 & 48.5 & 10.5 & 48.2 & 62.1 & 49.2 & 46.7 & 68.7 & 51.4 & 56.3 \\
InternVL3.5-8B~\cite{wang2025internvl3} & 59.7 & 72.7 & 48.4 & 51.2 & 58.9 & 72.2 & 47.4 & 72.1 & 62.4 & 47.6 & 27.4 & 62.5 & 43.4 & 52.5 \\
Qwen3-VL-8B~\cite{bai2025qwen3} & 56.7 & 72.4 & 49.3 & 54.1 & 60.4 & \underline{73.6} & 57.9 & \underline{73.5} & 67.6 & 53.0 & 32.2 & 72.9 & 51.8 & 59.0 \\

\rowcolor{DividerGray}
\multicolumn{15}{l}{\textit{Closed-source Models}} \\
Gemini-2.5-Flash~\cite{google2025gemini2.5} & \underline{68.3} & \underline{78.7} & 52.8 & \underline{57.6} & \underline{65.1} & \underline{73.6} & 57.9 & \underline{73.5} & 66.9 & 55.6 & 44.9 & 70.8 & 42.5 & 58.9 \\
GPT-5.2$^\star$~\cite{openai2026gpt52} & 64.0 & 73.0 & \textbf{57.7} & 57.4 & 63.5 & 71.5 & \textbf{73.7} & 71.5 & \underline{70.5} & \underline{61.9} & 46.4 & \textbf{78.8} & 56.4 & \underline{65.4} \\

\rowcolor{DividerGray}
\multicolumn{15}{l}{\textit{Spatially Specialized Models}} \\
SpaceQwen2.5-VL-3B~\cite{remyxai2025spaceqwen25vl} & 54.3 & 59.8 & 42.9 & 47.4 & 51.5 & 61.9 & 42.1 & 61.7 & 58.0 & 57.4 & \textbf{71.1} & 57.4 & \underline{58.6} & 58.7 \\
Spatial-MLLM-7B$^\ddagger$~\cite{wuspatial} & 52.9 & 49.3 & 42.7 & 46.6 & 47.5 & 50.4 & 36.8 & 50.3 & 37.3 & 29.2 & 33.4 & 37.5 & 29.4 & 33.3 \\
VLM3R-7B$^\ddagger$~\cite{fan2025vlm3rvisionlanguagemodelsaugmented} & 46.0 & 42.3 & 31.1 & 26.5 & 35.3 & 62.6 & 63.2 & 62.6 & 50.0 & 34.5 & 50.3 & 39.9 & 41.7 & 42.2 \\
SpaceMantis-8B~\cite{remyx2025spacemantis} & 41.4 & 57.1 & 42.9 & 47.2 & 48.9 & 62.3 & 52.6 & 62.3 & 45.1 & 38.2 & 57.5 & 42.3 & 41.3 & 42.7 \\
SpatialLLM-8B~\cite{ma2025spatialllm} & 50.6 & 76.7 & 49.5 & 48.5 & 58.4 & 57.8 & 52.6 & 57.8 & 49.7 & 53.5 & 44.3 & 48.3 & 38.1 & 48.8 \\
\midrule
\rowcolor{OursHighlight}
\textbf{Ours} & \textbf{68.6} & \textbf{84.8} & \underline{54.9} & \textbf{67.1} & \textbf{70.7} & \textbf{77.7} & \underline{68.4} & \textbf{77.6} & \textbf{72.0} & \textbf{63.0} & 45.5 & \underline{76.1} & \textbf{65.7} & \textbf{66.4} \\
\bottomrule
\end{tabular}
}
\end{table}

\subsection{Experimental Results}
\noindent\textbf{State-of-the-Art Performance Across Diverse Benchmarks. }
The proposed method consistently achieves the highest average accuracy across all three evaluation benchmarks, establishing a new state of the art for spatial reasoning tasks. 
As shown in \cref{tab:main}, the model significantly outperforms the strongest closed-source baseline, GPT-5.2, on both 3DSR ($70.7\%$ versus $63.5\%$) and SpatialScore ($77.6\%$ versus $71.5\%$). Furthermore, on the challenging Rel3D dataset, the approach yields an average score of $66.4\%$, surpassing GPT-5.2 by an absolute margin of $1.0\%$. These results highlight the effectiveness of the proposed design in resolving diverse spatial perception and reasoning challenges, demonstrating superiority over large-scale, general-purpose closed-source models and existing open-source solutions.

% 内容太多的话这一段可以不要
\noindent\textbf{Generalist Scalability Versus Domain-Specific Specialization. }
\cref{tab:main} reveals a notable trend regarding model architectures and pre-training paradigms: leading generalist large multimodal models currently surpass models that are explicitly designed or fine-tuned for spatial tasks. Closed-source generalist models dominate the baseline comparisons; for example, Gemini-2.5-Flash achieves an accuracy of $65.1\%$ on 3DSR, which substantially exceeds the performance of the top-performing spatially specialized model, SpatialLLM-8B ($58.4\%$). Even within the open-source community, generalist models demonstrate superior adaptability. For instance, Qwen3-VL-8B-Instruct exactly matches the performance of Gemini-2.5-Flash on the SpatialScore benchmark at $73.5\%$. While certain specialized models exhibit isolated strengths, such as SpaceQwen2.5-VL-3B achieving the highest accuracy in distance perception ($71.1\%$ on Rel3D Dis.), the overall average performance of these models remains constrained. This observation suggests that the extensive visual-linguistic alignment and world knowledge embedded in scaled generalist models currently provide a more robust foundation for complex spatial reasoning compared to narrow, domain-specific optimizations. Moreover, our method employs a \textit{training-free} paradigm, preserving the general knowledge of generalist models while establishing new state-of-the-art results for spatial reasoning tasks.

\setcounter{table}{1}
\setcounter{figure}{4}
\setcounter{equation}{9}

\begin{table}[t]
\centering
\caption{Performance of the base model, the proposed method, and the corresponding variants on the \textbf{3DSR}, \textbf{SpatialScore}, and \textbf{Rel3D} benchmarks, with the \textbf{View Reselect}, \textbf{Knowledge Base}, and \textbf{Coordinate Layout} modules individually ablated. The best and second-best results for each metric are indicated in \textbf{bold} and \underline{underlined}, respectively. All results represent accuracy (\%). The abbreviation w/o denotes ``without''.}
\label{tab:abla}
\resizebox{\linewidth}{!}{
\begin{tabular}{l|ccccc|ccc|cccccc}
\toprule
\rowcolor{LightBlue}
 & \multicolumn{5}{c}{\textbf{3DSR}} & \multicolumn{3}{c}{\textbf{SpatialScore}} & \multicolumn{6}{c}{\textbf{Rel3D}} \\
% \cmidrule(lr){2-6} \cmidrule(lr){7-9} \cmidrule(lr){10-15}
\rowcolor{LightBlue}
\multirow{-2}{*}{\textbf{Method}} & Hei. & Loc. & Ori. & MO. & \textbf{Avg.} & Ori. & Rea. & \textbf{Avg.} & Con. & Pos. & Dis. & Obs. & Ori. & \textbf{Avg.} \\
\midrule
Qwen3-VL-8B & 56.7 & 72.4 & 49.3 & 54.1 & 60.4 & 73.6 & 57.9 & 73.5 & 67.6 & 53.0 & 32.2 & 72.9 & 51.8 & 59.0 \\
\textbf{Ours} & \textbf{68.6} & \textbf{84.8} & \textbf{54.9} & \textbf{67.1} & \textbf{70.7} & \textbf{77.7} & \underline{68.4} & \textbf{77.6} & \textbf{72.0} & \textbf{63.0} & \textbf{45.5} & \textbf{76.1} & \textbf{65.7} & \textbf{66.4} \\
\midrule
w/o \textbf{View Reselect} & \underline{62.8} & \underline{80.7} &	\underline{52.4} &\underline{63.3} & \underline{66.8} &	74.6 &	\textbf{73.4} &	74.6 &	66.3 &	53.6 &	33.1 &	68.1 &	\underline{53.8} &	58.4 \\
w/o \textbf{Knowledge Base} & 54.1 & 77.4 & 51.7 &	53.4 &	61.2 &	\underline{77.3} &	63.2 &	\underline{77.2} &	\underline{69.4} &	\underline{58.0} &	\underline{41.3} &	\underline{75.8} &	50.8 &	\underline{62.1} \\
w/o \textbf{Coordinate Layout} & 59.9 & 73.7 &	51.2 &	56.2 &	61.5 &	75.5 &	63.2 &	75.5 &	67.1 &	54.4 &	35.5 &	68.3 &	52.2 &	58.9 \\
\bottomrule
\end{tabular}
}
\end{table}

\subsection{Ablation Study}
To verify the effectiveness of the proposed View Reselect, Knowledge Base, and Coordinate Layout modules, we conduct ablation experiments, as summarized in \cref{tab:abla}.

\noindent\textbf{Effect of View Reselect. }
Removing the View Reselect mechanism leads to consistent performance degradation across almost all tasks. Specifically, as shown in \cref{tab:abla}, the average accuracy drops from 70.7 to 66.8 on 3DSR, 77.6 to 74.6 on SpatialScore, and 66.4 to 58.4 on Rel3D, which even falls below the performance of the baseline. This phenomenon occurs because when the generated views are uninformative for visual question answering, the model tends to either ignore the visual inputs or formulate erroneous judgments based on misleading information. For SpatialScore, which lacks direct guidance from the Knowledge Base, the performance regresses nearly to the level of the baseline. Notably, on Rel3D, which contains numerous scenes with occlusions, the verification mechanism becomes critical; the absence of this mechanism causes the model to perform worse than the baseline. This highlights the essential role of the View Reselect module in facilitating robust spatial reasoning under ambiguous conditions.

\noindent\textbf{Effect of Knowledge Base. }
The Knowledge Base utilized in the proposed method is detailed in \cref{tab:knowledge_base}. Omitting this module results in severe performance drops, particularly on the 3DSR benchmark (decreasing from 70.7 to 61.2) and the Rel3D benchmark (decreasing from 66.4 to 62.1), as shown in \cref{tab:abla}. Because the Knowledge Base is explicitly tailored for 3D spatial reasoning tasks, removing it causes the view selection process of the model to become nearly random. This randomness leads to suboptimal visual inputs, thereby degrading performance across all benchmarks. These results confirm that the Knowledge Base provides crucial guidance for generating informative viewpoints.

\noindent\textbf{Effect of Coordinate Layout. }
To enable the model to derive accurate camera parameters from the description of the selected viewpoint, we incorporate a 3D coordinate system to facilitate precise spatial localization. In the ablation study, Step 2 is modified to allow the model to directly predict the yaw, pitch, and camera-to-scene distance required to transform a side view to the target viewpoint. The model subsequently computes the camera extrinsics from these predicted parameters and renders a novel view accordingly. The specific prompt utilized for this process is provided in \cref{sec:ablation_prompt}.

The results indicate that when the 3D Coordinate Layout is removed, the performance of the model degrades nearly to the level of the baseline. This degradation occurs because, without explicit 3D coordinate guidance, the model fails to construct a coherent spatial representation of the scene, rendering it unable to predict accurate camera parameters based on viewpoint descriptions. Furthermore, persistent hallucination issues are observed, such as confusion between positive and negative angles for left and right orientations. The Coordinate Layout enables the model to intuitively map semantic descriptions to precise camera poses while mitigating hallucinations, which is vital for reliable view synthesis.

\begin{figure}[t] 
\centering 
\includegraphics[width=\textwidth]{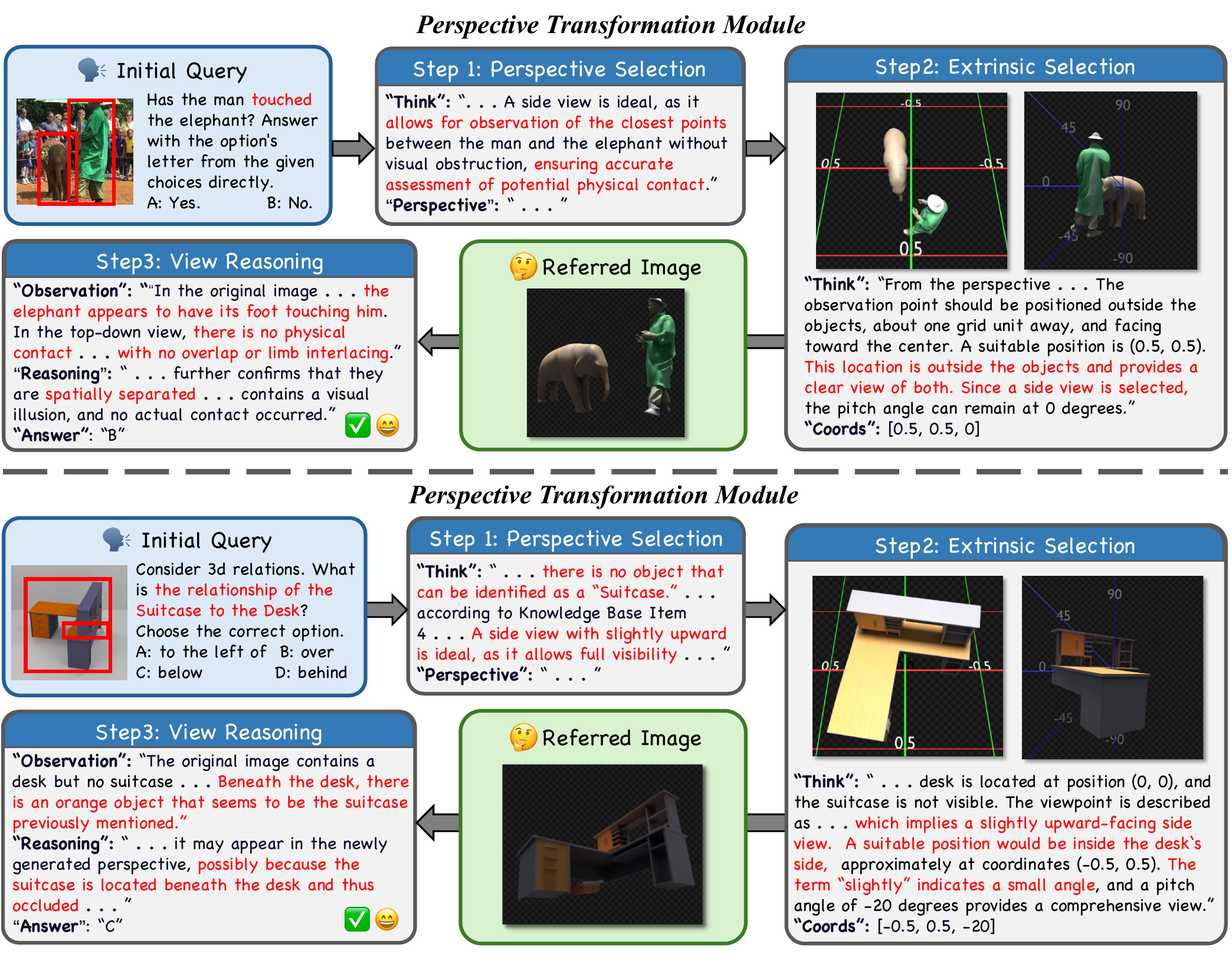}
\caption{The reasoning pipeline of the proposed method on two representative examples from the dataset. Each box presents the output information from the MLLM alongside the final answer to the question.} 
\label{fig:case11} 
\end{figure}

\subsection{Case Study}
\label{sec:case_study}

\noindent\textbf{Zero-Shot Reasoning. } The first example in \cref{fig:case11} corresponds to the Reach task within the SpatialScore dataset. The knowledge base contains no prior experience for solving this specific type of task. However, it encapsulates the utility of multiple viewpoints and the reasoning process required to identify the optimal perspectives for a given task. Coupled with the strong generalization capabilities of the MLLM, the model infers useful viewpoints to answer questions in novel scenarios by following the strategy for viewpoint selection from the knowledge base. Specifically, when determining whether two objects are in contact, a side view is sufficient to locate the closest points between them, thereby providing a favorable perspective for the MLLM to yield accurate judgments.

\noindent\textbf{Occluded View Reasoning. } The second example in \cref{fig:case11} belongs to the Position task within the Rel3D dataset, which features scenes with heavy occlusion. Based on the outputs of the MLLM across the entire pipeline, we observe that integrating 3D scene reconstruction enables the MLLM to actively switch viewpoints and observe the scene from an unobstructed perspective, thereby answering the question correctly. This outcome demonstrates the robust capability of the proposed method in reasoning about spatial relations involving partially occluded objects.

\noindent\textbf{Multi-Perspective Reasoning. } The examples in \cref{fig:case11} illustrate that during viewpoint selection, the model interacts continuously with the 3D scene to capture a set of multi-view images. Guided by carefully designed prompts, the model describes observations from each viewpoint and aggregates multi-view cues to draw accurate conclusions during the reasoning process. This approach leads to correct answers and effectively enhances the spatial understanding capabilities of the MLLM.

\section{Conclusion}
\label{sec:conclusion}
In this work, we propose a training-free framework designed to enhance the 3D spatial reasoning capabilities of Multimodal Large Language Models. By introducing a Visual Chain-of-Thought mechanism grounded in explicit 3D reconstruction, the proposed approach overcomes the limitations associated with the reliance on 2D visual priors. The framework dynamically synthesizes novel viewpoints through the iterative exploration of camera poses, emulating human spatial cognition to resolve visual ambiguities. Extensive evaluations demonstrate that the proposed method outperforms existing specialized models and general-purpose systems on multiple spatial reasoning benchmarks. Future research will explore the integration of this paradigm into dynamic environments to further advance the development of embodied intelligence.

\section*{Acknowledgements}

This work is supported in part by the National Natural Science Foundation of China (62192783, 62276128, 62406140), Young Elite Scientists Sponsorship Program by China Association for Science and Technology (2023QNRC001), the Key Research and Development Program of Jiangsu Province under Grant (BE2023019) and Jiangsu Natural Science Foundation under Grant (BK20221441, BK20241200). The authors would like to thank Huawei Ascend Cloud Ecological Development Project for the support of Ascend 910 processors.

\bibliographystyle{splncs04}
\bibliography{main}

\clearpage  % TODO FINAL: This \clearpage needs to be removed from both review and camera-ready versions.

\appendix
\section{Detailed Subtask Composition and Question Types of the Benchmarks}
\label{sec:appendix_task_details}

As discussed in \cref{sec:exp_setting}, the evaluation benchmarks utilized in this study consist of multiple subtasks, each encompassing a diverse set of specific question types and spatial relations. This section provides a comprehensive breakdown of the task configurations, specific question types, sample sizes, and relative proportions for the \textbf{3DSRBench}, \textbf{SpatialScore}, and \textbf{Rel3D} benchmarks, which are summarized in \cref{tab:benchmark_details}.

\noindent\textbf{The 3DSR Benchmark. }
The 3DSR benchmark used for evaluation contains 2,625 samples and is categorized into four distinct spatial reasoning subtasks:
\begin{itemize}
    \item \textbf{Location (Loc.) [33.34\%]:} This subtask evaluates the spatial placement of objects, encompassing the \texttt{\seqsplit{location\_next\_to}}, \texttt{\seqsplit{location\_closer\_to\_camera}}, and \texttt{\seqsplit{location\_above}} question types.
    \item \textbf{Multi-Object (MO.) [33.33\%]:} This subtask focuses on the relative configurations among multiple objects. The specific question types include \texttt{\seqsplit{multi\_object\_facing}}, \texttt{\seqsplit{multi\_object\_same\_direction}}, \texttt{\seqsplit{multi\_object\_parallel}}, \texttt{\seqsplit{multi\_object\_closer\_to}}, and \texttt{\seqsplit{multi\_object\_viewpoint\_towards\_object}}.
    \item \textbf{Orientation (Ori.) [20.00\%]:} This subtask focuses on the directional attributes of two objects, including \texttt{\seqsplit{orientation\_in\_front\_of}}, \texttt{\seqsplit{orientation\_viewpoint}}, and \texttt{\seqsplit{orientation\_on\_the\_left}}.
    \item \textbf{Height (Hei.) [13.33\%]:} This subtask assesses vertical spatial reasoning, specifically focusing on the \texttt{\seqsplit{Height\_Higher}} question type.
\end{itemize}

\noindent\textbf{The SpatialScore Benchmark. }
The SpatialScore benchmark selected for evaluation comprises 2,530 samples and focuses on orientation and reach relationships:
\begin{itemize}
    \item \textbf{Orientation (Ori.) [99.24\%]:} This subtask dominates the dataset, evaluating the \texttt{\seqsplit{3D Positional Relation}} type.
    \item \textbf{Reachability (Rea.) [0.76\%]:} This subtask represents a minor subset focused on \texttt{\seqsplit{Object Property}} type.
\end{itemize}

\begin{table}[t]
    \centering
    \caption{Knowledge Base for spatial and relational reasoning. Here, \textbf{Hei.} denotes \texttt{\protect\seqsplit{Height\_Higher}}, \textbf{M.O. 1} denotes \texttt{\protect\seqsplit{multi\_object\_viewpoint\_towards\_objec}}, \textbf{M.O. 2} denotes \texttt{\protect\seqsplit{multi\_object\_parallel}}, \textbf{Loc. 1} denotes \texttt{\protect\seqsplit{location\_closer\_to\_camera}}, \textbf{Loc. 2} denotes \texttt{\protect\seqsplit{location\_above}}, \textbf{Loc. 3} denotes \texttt{\protect\seqsplit{location\_next\_to}}, and \textbf{Task.} denotes Task Generalization.}
    \label{tab:knowledge_base}
    \renewcommand{\arraystretch}{1.5} % 增加行高，使长文本阅读更舒适
    \begin{tabularx}{\textwidth}{@{} c l X @{}}
        \toprule
        \rowcolor{LightBlue} \textbf{No.} & \textbf{Task} & \textbf{Strategy / Guideline} \\
        \midrule
        1 & \textbf{Hei.} 
          & When comparing heights, select a position where all objects are visible, view them at eye level, and compare which is the tallest. \\
          
        2 & \textbf{M.O. 1}
          & To determine which side an object is on (front, back, left, right), first identify the central object in question, then switch to a viewpoint behind it, i.e., transform into the coordinate system of that object. \\

        3 & \textbf{M.O. 2}
          & To judge whether objects are parallel, select a viewpoint that clearly shows the long edges of the objects, and check if the two long edges are parallel. \\
          
        4 & \textbf{Loc. 1}
          & When the question relates to the camera, the camera position should be at least the same as in the original image. \\
          
        5 & \textbf{Loc. 2}
          & To judge whether an object is directly above another, use a top-down view and project the objects onto the horizontal plane; an occlusion relationship indicates that one is directly above the other. \\
          
        6 & \textbf{Loc. 3}
          & To determine whether two objects are close, choose the viewpoint that reveals their maximum distance, then check if this maximum distance exceeds the diameter of the objects; if so, the objects may be considered close. \\
          
        7 & \textbf{Task.} 
          & Generalize to other task types based on the prior knowledge above. \\
        \bottomrule
    \end{tabularx}
\end{table}

\begin{table}[!t]
    \centering
    \caption{Detailed configurations of the question types and task compositions for the 3DSRBench, SpatialScore, and Rel3D benchmarks. The abbreviations follow the definitions provided in \cref{fig:benchmark_distribution}.}
    \label{tab:benchmark_details}
    \renewcommand{\arraystretch}{1.2}
    \resizebox{\linewidth}{!}{
    \begin{tabular}{@{}llcp{7.5cm}@{}}
        \toprule
        \rowcolor{LightBlue}
        \textbf{Benchmark} & \textbf{Subtask} & \textbf{Proportion} & \textbf{Specific Question Types / Task Composition} \\
        \midrule
        
        \textbf{3DSR} & Loc. & 33.34\% & \texttt{location\_next\_to}, \texttt{location\_closer\_to\_camera}, \texttt{location\_above} \\
        (N=2,625) & MO. & 33.33\% & \texttt{multi\_object\_same\_direction}, \texttt{multi\_object\_viewpoint\_towards\_object}, \texttt{multi\_object\_facing}, \texttt{multi\_object\_closer\_to}, \texttt{multi\_object\_parallel} \\
         & Ori. & 20.00\% & \texttt{orientation\_in\_front\_of}, \texttt{orientation\_viewpoint}, \texttt{orientation\_on\_the\_left} \\
         & Hei. & 13.33\% & \texttt{Height\_Higher} \\
        
        \midrule
        
        \textbf{SpatialScore} & Ori. & 99.24\% & \texttt{3D Positional Relation}, \texttt{Orientation} \\
        (N=2,530) & Rea. & 0.76\% & \texttt{Object Properties}, \texttt{Reachability} \\
        
        \midrule
        
        \textbf{Rel3D} & Pos. & 32.97\% & \texttt{over}, \texttt{below}, \texttt{under}, \texttt{to the side of}, \texttt{to the left of}, \texttt{to the right of}, \texttt{in front of}, \texttt{behind}, \texttt{around} \\
        (N=4,743) & Con. & 30.34\% & \texttt{on}, \texttt{on top of}, \texttt{in}, \texttt{inside}, \texttt{outside}, \texttt{touching}, \texttt{covering}, \texttt{leaning against}, \texttt{passing through} \\
         & Obs. & 15.90\% & \texttt{behind (w.r.t.\ you)}, \texttt{to the side of (w.r.t.\ you)}, \texttt{in front of (w.r.t.\ you)}, \texttt{to the left of (w.r.t.\ you)}, \texttt{to the right of (w.r.t.\ you)} \\
         & Ori. & 13.79\% & \texttt{faces away}, \texttt{points towards}, \texttt{points away}, \texttt{faces towards}, \texttt{aligned to} \\
         & Dis. & 7.00\% & \texttt{near}, \texttt{far from} \\
        
        \bottomrule
    \end{tabular}}
\end{table}

\noindent\textbf{The Rel3D Benchmark. }
The Rel3D benchmark is the largest dataset in this evaluation, containing 4,743 samples. It features a rich set of fine-grained 3D spatial relations distributed across five subtasks:
\begin{itemize}
    \item \textbf{Position (Pos.) [32.97\%]:} This subtask includes queries regarding relative positioning, specifically \texttt{\seqsplit{over}}, \texttt{\seqsplit{below}}, \texttt{\seqsplit{under}}, \texttt{\seqsplit{to the side of}}, \texttt{\seqsplit{to the left of}}, \texttt{\seqsplit{to the right of}}, \texttt{\seqsplit{in front of}}, \texttt{\seqsplit{behind}}, and \texttt{\seqsplit{around}}.
    \item \textbf{Containment (Con.) [30.34\%]:} This subtask evaluates contact and enclosing relations, including \texttt{\seqsplit{on}}, \texttt{\seqsplit{on top of}}, \texttt{\seqsplit{in}}, \texttt{\seqsplit{inside}}, \texttt{\seqsplit{outside}}, \texttt{\seqsplit{touching}}, \texttt{\seqsplit{covering}}, \texttt{\seqsplit{leaning against}}, and \texttt{\seqsplit{passing through}}.
    \item \textbf{Observer (Obs.) [15.90\%]:} This subtask focuses on egocentric or viewer-centric spatial relations, comprising \texttt{\seqsplit{behind} (w.r.t.\ you)}, \texttt{\seqsplit{to the side of} (w.r.t.\ you)}, \texttt{\seqsplit{in front of} (w.r.t.\ you)}, \texttt{\seqsplit{to the left of} (w.r.t.\ you)}, and \texttt{\seqsplit{to the right of} (w.r.t.\ you)}.
    \item \textbf{Orientation (Ori.) [13.79\%]:} This subtask assesses directional facing and alignment, encompassing \texttt{\seqsplit{faces away}}, \texttt{\seqsplit{points towards}}, \texttt{\seqsplit{points away}}, \texttt{\seqsplit{faces towards}}, and \texttt{\seqsplit{aligned to}}.
    \item \textbf{Distance (Dis.) [7.00\%]:} This subtask evaluates proximity relations, specifically \texttt{\seqsplit{near}} and \texttt{\seqsplit{far from}}.
\end{itemize}

\section{Prompt Details}
In this section, we detail the design of the prompts employed in the proposed method and the subsequent evaluation. For simplicity, the system prompt is set to \texttt{None}, with the exception of the perspective transformation module. The following subsections outline the specific prompts utilized for object extraction, the perspective transformation module, the baseline evaluation and the ablation study.

\subsection{Object Extraction Prompt Design}
As discussed in \cref{sec:3d_recon}, Qwen3-VL-235B-A22B is utilized to extract key object terms from the input image and the associated question. To ensure that the segmentation model receives comprehensive text queries for accurate mask generation, the MLLM is instructed to output a structured \texttt{JSON} format containing keywords at three levels of granularity: original keywords, abbreviated keywords, and expanded keywords. The exact prompt template employed for this process is provided below.

\begin{tcolorbox}[breakable,
    title=\textbf{Prompt 1: Object Extraction}, 
]
System Prompt
\noindent\textsc{System Prompt}\\
\texttt{None}

\vspace{0.8em} % 段落间距
% 2. User Prompt
\noindent\textsc{User Prompt}\\
You are required to generate a structured JSON output for the subsequent 3D reconstruction tasks, based on the provided original image and the related question. Please strictly adhere to the following rules:
\begin{enumerate}
    \item \textbf{Keywords}: Extract the core keywords directly from the question, retaining them in the original form.
    \item \textbf{Abbreviated keywords}: Abbreviate the phrases in the keywords, retaining only the core nouns.
    \item \textbf{Expanded keywords}: Based on an observation of the original image, generate distinct expanded descriptions from the following three perspectives:
    \begin{itemize}
        \item \textbf{Positional relationship}: Describe the position of the object in the image (\eg, left, center, top) and the relative position to other objects or people in the scene (\eg, in front of . . . , to the right of . . . , occluded by . . . ).
        \item \textbf{Appearance features}: Supplement the visual details such as the color, shape, material, and components of the object.
        \item \textbf{Synonym relationship}: Use alternative synonyms or related concepts to describe the objects identified in the keywords.
    \end{itemize}
\end{enumerate}

Output format is as follows:
\begin{verbatim}
{
    "Keywords": [..., ..., ...],
    "Abbreviated keywords": [..., ..., ...],
    "Expanded keywords": [..., ..., ...]
} 
\end{verbatim}

\end{tcolorbox}

\subsection{Perspective Transformation Prompt Design}
\noindent\textbf{Step 1: Viewpoint Selection.}
In this initial step, the objective is to prompt the MLLM to determine the most advantageous viewpoint for solving the given spatial reasoning question. As illustrated in the prompt template below, the model is provided with the original image, the specific question, and the relevant items retrieved from the external knowledge base. To ensure a logical decision-making process, a \texttt{JSON} output format is enforced. The model is instructed to first articulate the reasoning process in the \texttt{"Think"} field, utilizing either the provided knowledge items or the inherent spatial imagination capabilities of the model. Subsequently, the model must output a concise linguistic description of the optimal viewing angle in the \texttt{"Perspective"} field, which serves as the target textual description for the subsequent prediction step of the camera extrinsic matrices.

\begin{tcolorbox}[breakable,
    title=\textbf{Prompt 2: Viewpoint Selection}, 
]
% 1. System Prompt
\noindent\textsc{System Prompt}\\
Assume you are an expert with 3D spatial reasoning capabilities.

Your final goal is to correctly answer questions involving spatial understanding.

To help you solve these problems, I have broken the solution process into several steps.

Please follow my instructions strictly step by step, and put your answers inside the specified tags according to the required format.

\vspace{0.8em} % 段落间距
% 2. User Prompt
\noindent\textsc{User Prompt}\\
\noindent\textbf{Input above:} An original image, along with a question regarding the spatial understanding of the image.

\vspace{0.5em}
\noindent\textbf{Requirement:} Based on the provided knowledge from the experience library and independent judgment, select the viewpoint position that is most conducive to answering the question. 
The viewpoint position refers to the space in all directions around the center, with the scene in the image acting as the observation center. \\
Examples: behind a specific object, a position where the objects relevant to the question can be observed simultaneously, a top-down view of the scene, \etc.

\vspace{0.5em}
\noindent\textbf{Knowledge from the library:}\\
\{\}

\vspace{0.5em}
\noindent\textbf{Output json format requirement:}
\begin{verbatim}
{
    "Think": ...,
    "Perspective": ...
}
\end{verbatim}

\noindent Where:
\begin{enumerate}
    \item The \textbf{"Think"} tag should contain the reasoning for selecting the viewpoint position. If similar problems and the corresponding viewpoint selections can be found in the experience library, please refer to them; otherwise, based on independent understanding and imagination, analyze the scene expected from different viewpoints and determine whether this scene can directly answer the question, thereby selecting the optimal viewpoint position.
    \item The \textbf{"Perspective"} tag should contain a brief description of the selected viewpoint position.
\end{enumerate}

\end{tcolorbox}

\noindent\textbf{Step 2: Coordinate Selection.}
Building upon the target perspective described in the previous step, the MLLM is tasked with predicting the corresponding spatial coordinates and the viewing angle to compute the camera extrinsic. To visually ground this numerical prediction, the MLLM is provided with two rendered auxiliary images: a top-down map of the $xy$-plane and a side-view map illustrating the vertical pitch angles. The model is instructed to first locate the relevant objects within these coordinate systems and subsequently infer the discrete position of the camera $(\hat{x}', \hat{y}')$ and the pitch angle $\widehat{pitch}$. These predicted values are then converted into continuous camera extrinsic matrices, as detailed in the main text.

\begin{tcolorbox}[breakable,
    title=\textbf{Prompt 3: Coordinate Selection}, 
]
% 1. System Prompt
\noindent\textsc{System Prompt}\\
Assume you are an expert with 3D spatial reasoning capabilities.

Your final goal is to correctly answer questions involving spatial understanding.

To help you solve these problems, I have broken the solution process into several steps.

Please follow my instructions strictly step by step, and put your answers inside the specified tags according to the required format.

\vspace{0.8em} % 段落间距
% 2. User Prompt
\noindent\textsc{User Prompt}

\noindent\textbf{Input above:} The first image is a top-down XY plane coordinate map of the 3D scene for the objects relevant to the question, which are extracted from the original image. The red line represents the X-axis, the green line represents the Y-axis, and the central position is the point (0, 0). The interval between any two parallel lines is 0.5. The white numbers on the principal axes indicate the values of the coordinates represented by the corresponding lines. The second image is a side view of the scene, displaying the vertical plane of the 3D scene. The end of the blue line marks the position of the camera, and the blue line represents the viewing line of the camera (\ie, from the end to the center). Different blue lines correspond to different angle values (pitch angles). The white numbers on the blue lines indicate the pitch angle represented by the respective line.

\vspace{0.5em}
\noindent\textbf{Requirements:} Based on the perspective position description output in the previous round (\ie, \{\}) and the evaluation of the new coordinate map, select the coordinates that match the described position, including the XY plane coordinates and the pitch angle, in order to look toward the center from these coordinates.

\vspace{0.5em}
\noindent\textbf{Notes:}
\begin{enumerate}
    \item The selected XY plane coordinates should not be located inside the object (\eg, [0, 0]), but rather outside the object at a distance of approximately one empty grid interval.
    \item The value range of the pitch angle is [-90, 90].
\end{enumerate}

\vspace{0.5em}
\noindent\textbf{Output json format requirements:}
\begin{verbatim}
{
    "Think": ...,
    "Coords": [x, y, pitch]
}
\end{verbatim}

\noindent Where:
\begin{enumerate}
    \item The \textbf{"Think"} tag must include the sequential observation results of the plane coordinate map and the side view (\ie, state the coordinate position of each object). Select the coordinates in the plane map and the side view that match the description of the perspective position and the sequential requirements based on the observation results.
    \item The \textbf{"Coords"} tag must contain the selected coordinate position.
\end{enumerate}

\end{tcolorbox}

\noindent\textbf{Step 3: Final Reasoning and Answer.}
In this final step, the pipeline renders a novel view using the camera extrinsic predicted in Step 2. The MLLM is then provided with this synthesized image, acting as both a view evaluator and a spatial reasoner. Initially, the model must verify the quality of the rendered view against strict criteria, including object completeness, appropriate scaling, and consistency with the initial perspective description. If the synthesized view exhibits critical defects, the MLLM is instructed to output \texttt{"None"}, which triggers an iterative refinement process to regenerate the camera parameters (as detailed in \cref{sec:persp_trans}). Once the synthesized view is validated, the MLLM seamlessly integrates the visual cues from the original image, the coordinate maps, and the novel view to perform comprehensive spatial reasoning and deduce the final answer.

\begin{tcolorbox}[breakable,
    title=\textbf{Prompt 4: Final Reasoning and Answer}, 
]
% 1. System Prompt
\noindent\textsc{System Prompt}\\
Assume you are an expert with 3D spatial reasoning capabilities.

Your final goal is to correctly answer questions involving spatial understanding.

To help you solve these problems, I have broken the solution process into several steps.

Please follow my instructions strictly step by step, and put your answers inside the specified tags according to the required format.

\vspace{0.8em} % 段落间距
% 2. User Prompt
\noindent\textsc{User Prompt}\\
\noindent\textbf{Input above:} The last image is a perspective transformation image captured from the selected perspective position.

\vspace{0.5em}
\noindent\textbf{Requirements:} Based on all the obtained images, provide the observations and the reasoning analysis for the question regarding the original image, and finally derive the answer.

\vspace{0.5em}
\noindent\textbf{Notes:} Please analyze the final perspective transformation image and judge whether the following conditions are met in sequence. If any condition is not met, output the answer as "None":
\begin{enumerate}
    \item All objects related to the question are present in the frame;
    \item The displayed perspective is consistent with the perspective description obtained in Step 1 (\ie, \{\});
    \item There is no issue of incomplete display or overly small display regarding the objects. (Partial occlusion between objects is permitted.)
\end{enumerate}

\vspace{0.5em}
\noindent\textbf{Output json format requirements:}
\begin{verbatim}
{
    "Observation": ...,
    "Reasoning": ...,
    "Answer": ...
}
\end{verbatim}

\noindent Where:
\begin{enumerate}
    \item The \textbf{"Observation"} tag includes the observations of the original image, the plane coordinate map, the side view, and the perspective transformation image, providing a visual description related to the question for each image respectively.
    \item The \textbf{"Reasoning"} tag includes the solution process for the question based on the observations, analyzes whether the perspective images meet all the aforementioned requirements and conditions, and finally infers the answer to the question.
    \item The \textbf{"Answer"} tag contains the final answer to the question; output only "None" if any of the aforementioned conditions are not met upon analysis.
\end{enumerate}

\end{tcolorbox}

\subsection{Evaluation Prompt Design}
This subsection describes the design of the prompt used for the evaluations presented in \cref{tab:main}. To further evaluate the instruction-following capability of both the \textit{general-purpose models} and the \textit{spatially specialized models}, we require the models to generate responses in a valid \texttt{JSON} format. Any deviation from this specified format is treated as an incorrect response.

\begin{tcolorbox}[breakable,
    title=\textbf{Prompt 5: Evaluation of Baselines},
]
% 1. System Prompt
\noindent\textsc{System Prompt}\\
\texttt{None}

\vspace{0.8em} % 段落间距
% 2. User Prompt
\noindent\textsc{User Prompt}\\
Please answer the question based on the image provided. You are required to generate a structured JSON output. Please strictly follow the rules below:

\begin{enumerate}
    \item \textbf{Reasoning}: Briefly explain your reasoning.
    \item \textbf{Answer}: Output your final single choice.
\end{enumerate}

Output format is as follows:
\begin{verbatim}
{
    "Reasoning": “...",
    "Answer": "A"
}
\end{verbatim}

\end{tcolorbox}

\subsection{Alternative Viewpoint Parameter Prediction Prompt Design}
\label{sec:ablation_prompt}
As an alternative to the coordinate selection method, this prompt variant tasks the MLLM with directly predicting the continuous  transformation parameters (yaw, pitch, and distance) required to transition from an initial side-view perspective to the target viewpoint. The prompt is provided below.
\begin{tcolorbox}[breakable,
    title=\textbf{Prompt 6: Direct Viewpoint Transformation Prediction}, 
]
% 1. System Prompt
\noindent\textsc{System Prompt}\\
Assume you are an expert with 3D spatial reasoning capabilities.

Your final goal is to correctly answer questions involving spatial understanding.

To help you solve these problems, I have broken the solution process into several steps.

Please follow my instructions strictly step by step, and put your answers inside the specified tags according to the required format.

\vspace{0.8em} % 段落间距
% 2. User Prompt
\noindent\textsc{User Prompt}\\
\noindent\textbf{Input:} \\
\texttt{[Image: Side View]} \\
The provided image is a side view illustrating the vertical plane of the 3D scene.

\vspace{0.5em}
\noindent\textbf{Requirement:} \\
Using this side view as the initial reference perspective, determine the necessary viewpoint transformation parameters required to transition to the optimal target perspective previously described in Step 1 (\ie, \texttt{\{Target\_Perspective\}}).

\vspace{0.5em}
\noindent\textbf{Notes (Coordinate System \& Parameters):} \\
Assume the scene is abstracted into a 3D coordinate system, where the $xy$-plane represents the ground level and the $z$-axis represents height. The camera viewpoint is constrained within a spherical coordinate space centered on the target object. 
Consequently, the viewpoint transformation is defined by three parameters: \texttt{yaw}, \texttt{pitch}, and \texttt{distance}. \\
The detailed specifications for these parameters are as follows:
\begin{enumerate}
    \item \textbf{yaw}: A continuous value in the range $[-180, 180]$. A negative degree indicates a leftward rotation around the target object from the current observation position, while a positive degree indicates a rightward rotation.
    \item \textbf{pitch}: A continuous value in the range $[-90, 90]$. A positive value indicates looking down at the target object from above, while a negative value indicates looking up from below.
    \item \textbf{distance}: A floating-point number with one decimal place in the range $[0.0, 10.0]$. A larger value pulls the camera further away from the object, while a smaller value brings it closer.
\end{enumerate}

\vspace{0.5em}
\noindent\textbf{Output Format:} \\
Please output your response strictly in the following JSON format:
\begin{verbatim}
{
    "View_Analysis": "...",
    "View_Params": [yaw, pitch, distance]
}
\end{verbatim}

\noindent \textbf{Where:}
\begin{enumerate}
    \item \texttt{"View\_Analysis"}: This field should contain your step-by-step spatial reasoning for calculating the parameter transformations needed to shift from the current viewpoint to the optimal target viewpoint.
    \item \texttt{"View\_Params"}: This field should contain a list of the three calculated viewpoint transformation parameters to achieve the target perspective from Step 1. \textit{Note: Assume the camera's initial position in the provided side view corresponds to the default parameter state: \texttt{[0, 0, 3]}.}
\end{enumerate}

\end{tcolorbox}

\end{document}